%% file: main.tex
\documentclass[runningheads]{llncs}
\usepackage[T1]{fontenc}
\usepackage{hyperref}
\usepackage{graphicx,verbatim}
\usepackage{subcaption}    
\usepackage{cleveref}
\usepackage{enumitem}
\usepackage{booktabs}
\usepackage{multirow}
\usepackage{pifont} 
\usepackage[table]{xcolor}
\usepackage{tabularx}
\usepackage{tikz}
\usepackage{bm}
\usepackage{hhline}

\definecolor{upgreen}{HTML}{008000}
\newcommand{\up}[1]{\textcolor{upgreen}{\scriptsize $\uparrow$#1}}
\newcommand{\down}[1]{\textcolor{red}{\scriptsize $\downarrow$#1}}
\begin{document}
\title{Curia-2: Scaling Self-Supervised Learning for Radiology Foundation Models}
%
\author{Antoine Saporta\thanks{Equal contribution.}\inst{1} \and
Baptiste Callard\protect\footnotemark[1]\inst{1} \and Corentin Dancette\inst{1} \and Julien Khlaut\inst{1,2,3} \and Charles Corbière\inst{1} \and Leo Butsanets\inst{1} \and Amaury Prat\inst{1} \and Pierre Manceron\inst{1}}
\authorrunning{A. Saporta et al.}
%
\institute{Raidium, 27 rue du Faubourg Saint-Jacques, 75014 Paris, France.  \\ 
\email{firstname.lastname@raidium.eu}
\and Department of Vascular and Oncological Interventional Radiology, Hôpital Européen Georges Pompidou, AP-HP, Paris, France
\and Faculté de Santé, Université Paris-Cité, Paris, France.}


  
\maketitle              

\begin{abstract}
The rapid growth of medical imaging has fueled the development of Foundation Models (FMs) to reduce the growing, unsustainable workload on radiologists.
While recent FMs have shown the power of large-scale pre-training to CT and MRI analysis, there remains significant room to optimize how these models learn from complex radiological volumes. Building upon the Curia framework, this work introduces Curia-2, which significantly improves the original pre-training strategy and representation quality to better capture the specificities of radiological data. The proposed methodology enables scaling the architecture up to billion-parameter Vision Transformers, marking a first for multi-modal CT and MRI FMs. Furthermore, we formalize the evaluation of these models by extending and restructuring CuriaBench into two distinct tracks: a 2D track tailored for slice-based vision models and a 3D track for volumetric benchmarking. Our results demonstrate that Curia-2 outperforms all FMs on vision-focused tasks and fairs competitively to vision-language models on clinically complex tasks such as finding detection. Weights will be made publicly available to foster further research.
\keywords{Foundation Models \and Computational Radiology.}

\end{abstract}
\section{Introduction}
Radiology is the cornerstone of modern diagnostic medicine, providing the essential visual data required to detect disease, monitor treatment efficacy, and guide surgical interventions across nearly every clinical specialty. 

Foundation Models (FMs) represent a paradigm shift in radiological AI, moving away from task-specific architectures toward broad representations learned from unlabeled data. In the natural vision domain, self-supervised models such as DINO~\cite{oquab2023dinov2,simeoni2025dinov3} have demonstrated that large-scale pre-training enables high performance on downstream tasks with minimal fine-tuning. 
Yet, adapting these methods requires handling the uniqueness of radiological data.

Recent advancements have begun to address these challenges, with numerous medical foundation models offering diverse solutions, including MedImageInsight~\cite{codella2024MedImageInsight}, BioMedCLIP~\cite{zhang2025BioMedCLIP}, MedGemma~\cite{sellergren2025medgemma}, Merlin~\cite{blankemeier2024merlin}, Pillar-0~\cite{agrawal2025pillar}, or Curia~\cite{dancette2025curia}. Among these, Curia leverages self-supervised pretraining on over 200 million CT and MR images and introducing CuriaBench, a 19-task benchmark designed to evaluate clinical versatility on these modalities. While Curia demonstrated the power of scale, the efficiency of the training strategy remains a challenge due to DINOv2 being fundamentally tailored for natural images. 

In this work, we introduce several optimizations to refine Curia's pre-training strategy. More specifically, we propose multiple adaptions to the data augmentations and objective functions of DINOv2 to explicitly address the inherent characteristics of radiological data.
Leveraging the efficiency and stability provided by our method, we introduce a new family of models, Curia-2, which scales up to billion-parameter architectures, a first for multi-modal CT and MRI FMs. This family consists of Curia-2 B (86M), Curia-2 L (303M), and Curia-2 g (1.3B).  To ensure a rigorous assessment of these advancements, we reformulate CuriaBench into two distinct tracks: a 2D track dedicated to the evaluation of slice-based FMs, and a 3D track providing a standardized benchmark for all radiological FMs. Our results show that the Curia-2 family outperforms all existing slice-based FMs across both tracks and performs competitively with 3D-native FMs on the 3D track.

\input{figures/few-shot-ablations-v2}

\section{Method}

\subsection{Large-scale Pre-training}
Our method builds on Curia~\cite{dancette2025curia}, which adapts the DINOv2~\cite{oquab2023dinov2} framework to large-scale radiology data. DINOv2 algorithm combines two objectives: (1) DINO loss, which aligns global representations between a student and a momentum teacher via a cross-entropy self-distillation, (2) the iBOT loss, which enforces consistency between masked patch-level representations across views. Together, these losses enable learning both global and local visual representations from unlabeled data.
This section outlines the methodological differences between our approach and both Curia and DINOv2, focusing on improved scalability for radiological data. These fundamental modifications significantly alter the training dynamics and the behavior of representation learning.

\paragraph{Low Pre-training Resolution.}
In contrast to Curia, which trains on 512-resolution square images, we train at 256 resolution. This configuration significantly lowers the computational cost per sample, enabling more optimization steps for an equivalent compute budget and improving scalability. For context, DINOv2 and its successor DINOv3~\cite{simeoni2025dinov3} are trained at resolutions of 224 and 256, respectively.

\paragraph{Content-Aware Cropping.}
DINOv2 generates multiple random crops of the input image, on which student and teacher representations are aligned using the DINO objective. While uniform random cropping is well-suited to natural images, it is less appropriate for radiological data, where a substantial portion of images consists of non-informative background. To ensure that feature alignment is conducted on meaningful content, we introduce a two-stage content-aware filtering during crop generation: First, input images that contain more than 50\% uniform background are discarded. In the remaining samples, we generate global and local crops containing at most 70\% background.
This mechanism significantly improves DINOv2's stability on radiological images, especially in low-resolution training, where uninformative background crops are more prevalent.

\input{figures/masking-v2}
\paragraph{Anatomically-Guided Masking.}
The blockwise masking with a uniform prior used in iBOT~\cite{zhou2021ibot} suffers from similar limitations when applied to radiological images. While the iBOT loss requires the model to reconstruct features of masked patches, a large fraction of the input typically consists of non-informative background; consequently, the uniform prior frequently targets these empty regions, rendering the task trivially simple. We improve upon this strategy by implementing a Gaussian-weighted prior centered on anatomical regions. To maintain spatial consistency, we adjust the Gaussian centers to account for the offsets introduced by global cropping. This biases the masking process toward informative structures, illustrated in Fig.~\ref{fig:masking}, generating a more meaningful learning signal.

\paragraph{Clinical-Prior Regularization.}
KoLeo loss is the regularizer employed in DINOv2 to encourage a uniform span of features within a batch. Because KoLeo indiscriminately pushes nearest neighbors apart to satisfy an entropy requirement, it may force the model to artificially separate representations that are clinically similar, making it ill-suited for radiological images. Instead, we opt for the SigReg loss~\cite{balestriero2025lejepa}. SigReg regularizes the global latent geometry by pulling the distribution toward an isotropic Gaussian. In the context of radiology, where the anatomical topology across different patients follows a highly consistent structural prior, it allows these points to remain naturally clustered while ensuring the overall embedding space remains dense. SigReg preserves the nuances necessary for clinical differentiation without inflating the variance of similar anatomical structures.

\paragraph{High-Resolution Fine-Tuning.}
To maintain comparability with Curia while pre-training at a lower resolution, we incorporated a final training stage to scale the input resolution to 512. We first evaluated a direct upscaling approach by interpolating the ViT’s positional embeddings. This straightforward adaptation yielded performance gains across most tasks compared to the 256 baseline. To further optimize high-resolution performance, we conducted a brief fine-tuning phase at resolution 512, following established protocols~\cite{oquab2023dinov2,simeoni2025dinov3}. This stage requires significantly fewer training steps and incurs minimal computational overhead, thereby preserving the efficiency gains of our low-resolution pre-training.

\subsection{Data Preparation}
The pre-training dataset, shared with Curia, consists of routine cross-sectional clinical examinations acquired from 2019 to 2022 through the collaboration with a private hospital. All data were fully anonymized, including removal of identifying metadata and defacing of head-containing scans. The initial dataset included 130 TB of data, with 164 million CT images and 64 million MR images. For quality control, only 3D CT and MR examinations with at least five images were retained, while low-quality localizer or scout sequences were excluded.

\subsection{Training Details}
We trained three variants of ViT architectures~\cite{vit}: ViT-B (86M), ViT-L (303M), and ViT-g (1,3B). All model configurations were scaled to train on 400M images: Curia-2 B was trained for 125,000 iterations across 32 NVIDIA A100 (64GB) GPUs with a batch size of 120, with 2 global crops and 8 local crops generated for each image in the batch; Curia-2 L for 156,300 iterations on 64 GPUs with a batch size of 48; and Curia-2 g for 156,300 iterations on 128 GPUs with a batch size of 20. High-resolution fine-tuning stages were run for 20,000 iterations across 8 GPUs, with batch sizes adjusted to hardware constraints. Total compute for the final Curia-2 models amounted to approximately 1,820, 3,270, and 9,680 GPU hours for the B, L, and g variants, respectively.

\section{Experiments}
\subsection{Evaluation Protocol}
All models were evaluated based on the protocol defined in Curia~\cite{dancette2025curia}. For each downstream task, we trained a simple classification, regression, or survival head on top of the model, without fine-tuning its encoder weights. For every model, we report the average performance over 5 runs. This strategy enables a lightweight, fair comparison of multiple foundation models.

\input{datasets/table}

We constructed our evaluation benchmark following Curia\-Bench~\cite{dancette2025curia} tasks and datasets for both ablation studies and comparative evaluations, described in Table~\ref{tab:tasks}. To ensure a rigorous comparison, the benchmark incorporates two primary methodological principles: (1) evaluation is conducted across two parallel tracks, a 2D track and a 3D track, where each task may reside in either or both, depending on the dataset structure. The 2D track is dedicated to 2D FMs, while the 3D track serves as a universal evaluation benchmark across all models -- for slice-based models, features are averaged on the volume for all 3D tasks. 
(2) Ten tasks from the 2D track are designated for the development and ablation of Curia-2, while the remaining tasks are reserved to evaluate the generalization of our learned representations. These tasks were selected based on their discriminative power as reported in the CuriaBench study, while ensuring diverse coverage of task categories, modalities, and anatomical regions. Furthermore, we incorporate a multi-finding screening task as proposed by Pillar-0~\cite{agrawal2025pillar}.

\subsection{Ablation Study}
\label{sec:ablation}
We conduct an incremental ablation study to evaluate our methodology, beginning with a standard DINOv2 baseline using a ViT-B backbone at 256 resolution. Note this differs a bit from Curia B, which was trained at 512 resolution.


As shown in Table~\ref{tab:ablation} and in Fig.~\ref{fig:sl_fs}a, the ViT-B model initially suffers from training collapse. Our cropping strategy, requiring crops to contain at least 30\% foreground signal, is critical for stability (+34.6 points). 
Since Curia B did not report such instability, we hypothesize that working with crops of half the resolution inherently leads to a higher frequency of empty samples.


\noindent Our results confirm that the standard KoLeo loss is ill-suited for the structural homogeneity of medical data, as removing this local distance penalty yields a +1.6 points increase. 
We instead adopt SigReg, which promotes a dense embedding space without sacrificing performance.

\noindent Replacing uniform masking with Gaussian sampling further refines the learning signal (+0.2 points). This focuses the iBOT reconstruction task on the central anatomical regions where clinical signal is most concentrated. 

\noindent Furthermore, we found that increasing the learning rate to $10^{-3}$ from the original $5\cdot10^{-4}$ further optimizes training dynamics (+0.5 points), leveraging the increased stability afforded by our architectural and sampling modifications.

Finally, we evaluated the staged transition to 512 resolution:
Positional embedding interpolation alone outperforms evaluation at the native 256 resolution (+1.1 points), high-resolution fine-tuning further refining the features (+0.2). 

As shown in the final rows of Table~\ref{tab:ablation}, scaling model size to ViT-L and g yields substantial improvements, reaching 88.1\% and 88.7\%, respectively, confirming that our modifications provide a robust foundation for larger-scale ViTs.
\input{tables/ablation}
\input{figures/similarity-v3}

\subsection{Results on the 2D and 3D Tracks}
\input{tables/2dresults}
\input{tables/3dresults}

On the 2D track (Table~\ref{tab:2dresults}), Curia-2 L and Curia-2 g set a new state-of-the-art with an average of 88.5\%, significantly exceeding competing models and demonstrating superior robustness across all categories. 
MedImageInsight~\cite{codella2024MedImageInsight}, best non-Curia model, notably performs well in specific emergency tasks (E2, E3) but falters in anatomical tasks. Curia-2 consistently improves upon its predecessor (+0.6\% for ViT-B and +0.9\% for ViT-L). Notably, while Curia was already boasting high performance on anatomical tasks (A1-3), the Curia-2 g advances anatomical understanding with peak performance in all of them, a capability further evidenced by the robust, cross-modal alignments shown in Fig.~\ref{fig:sim_maps}.
In particular, in a few-shot scenario (Fig.~\ref{fig:sl_fs}b), Curia-2 achieves equivalent performance milestones much earlier in the training process than competing models, highlighting its superior data efficiency.

On the 3D track (Table~\ref{tab:3dresults}), the Curia-2 family outperforms all competing FMs, with Curia-2 L securing the top overall position at an 88.6\% average, while Curia-2 g achieves remarkable performance on anatomy-related tasks (A1-3). Notably, this represents a substantial leap from its predecessor, improving upon Curia L by +4.1\%.
Finally, while findings detection (F) is traditionally thought to require the domain-specific vision-language knowledge of architectures like Merlin~\cite{blankemeier2024merlin} and Pillar-0~\cite{agrawal2025pillar}, the slice-based Curia-2 models remain competitive on this task with a peak AUC of 80.2\% compared to Pillar-0's 82.4\%.  

\section{Conclusion}

In this work, we introduced a refined pre-training recipe to account for the unique characteristics of radiological imaging. The proposed method, leading to Curia-2, stabilizes training and achieves superior performance over the original Curia. 
Notably, Curia-2 demonstrates a consistent scaling benefit from ViT-B to ViT-L, a trend that remained inconclusive in Curia. Yet, the performance of Curia-2 g remains close to that of Curia-2 L, indicating that the challenge of scaling to billion-parameter models in medical imaging is not yet fully resolved. Curia-2 establishes a new state-of-the-art in vision-focused radiological tasks and bridges the gap with vision-language models on complex tasks like findings detection.

Future work will explore ultra-large-scale strategies from the natural image community, such as drastic increases in data, batch size and training duration, alongside post-training refinements like Gram anchoring from DINOv3. By releasing our weights as open-source, we aim to provide a robust backbone for the community and accelerate research in general-purpose computational radiology.

    

\begin{credits}
\textbf{\ackname} We acknowledge the EuroHPC Joint Undertaking for awarding this project access to the EuroHPC supercomputer LEONARDO, hosted by CINECA (Italy) and the LEONARDO consortium through an EuroHPC AI Factory Access call.
\end{credits}

%
%
%
\bibliographystyle{splncs04}
\bibliography{main, raidium}

\end{document}

%% file: figures/few-shot-ablations-v2.tex

\begin{figure}[!b]
    \centering
    
    \begin{tikzpicture}
        \node[anchor=south west, inner sep=0] (image) at (0,0) {\includegraphics[width=0.496\textwidth]{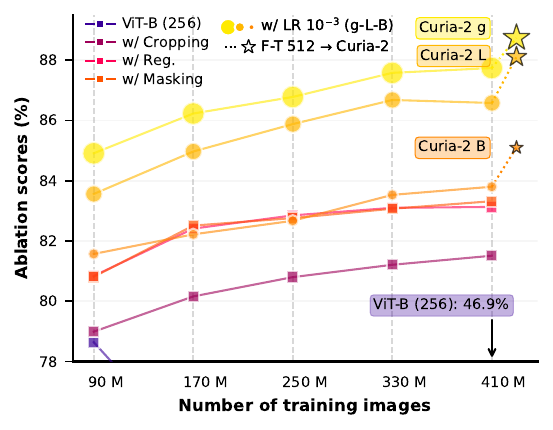}};
        \begin{scope}[x={(image.south east)},y={(image.north west)}]
            \node[anchor=north west, font=\bfseries] at (0.02,0.10) {(a)};
        \end{scope}
    \end{tikzpicture}
    \begin{tikzpicture}
        \node[anchor=south west, inner sep=0] (image) at (0,0) {\includegraphics[width=0.496\textwidth]{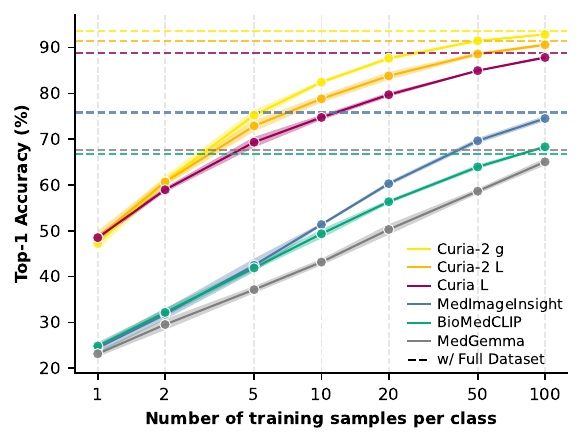}};
        \begin{scope}[x={(image.south east)},y={(image.north west)}]
            \node[anchor=north west, font=\bfseries] at (0.02,0.10) {(b)};
        \end{scope}
    \end{tikzpicture}
    \caption{(a) \textbf{Scaling Laws.} The training dynamic steadily improves after each modification (cf. Section~\ref{sec:ablation}). \ding{73}: fine-tuning in 512 resolution. (b) \textbf{Few-shot Learning (A1).} Curia-2 g demonstrates quicker convergence and requires significantly fewer samples to achieve the same performance as other models.}
    \label{fig:sl_fs}
\end{figure}

%% file: figures/masking-v2.tex
\begin{figure}[b]
    \centering
    \begin{minipage}[t]{0.32\textwidth}
        \centering
        \includegraphics[width=0.48\linewidth]{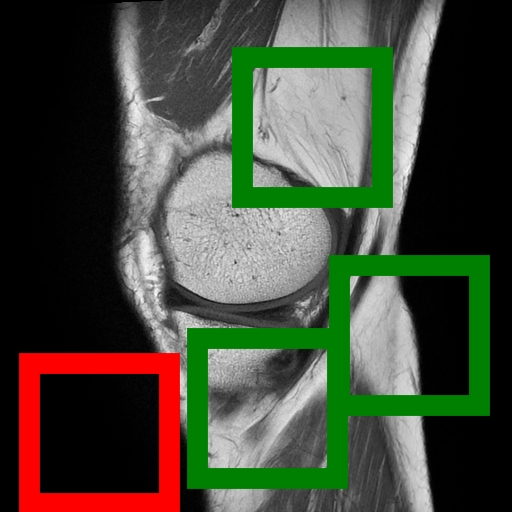}\hfill
        \includegraphics[width=0.48\linewidth]{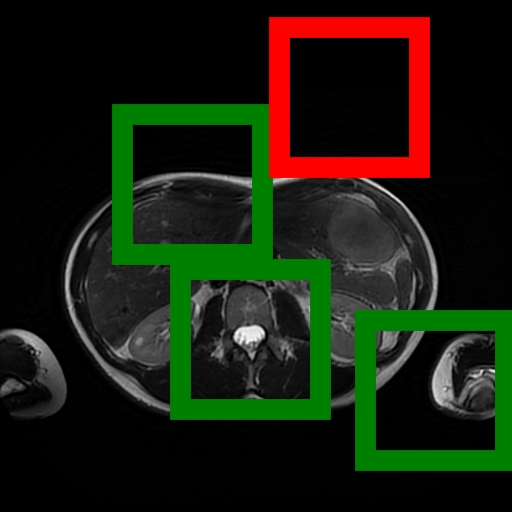}\\[4pt]
        \small Content-Aware Cropping
    \end{minipage}%
    \hfill
    \vrule width 0.5pt 
    \hfill
    \begin{minipage}[t]{0.65\textwidth}
        \centering
        \includegraphics[width=0.24\linewidth]{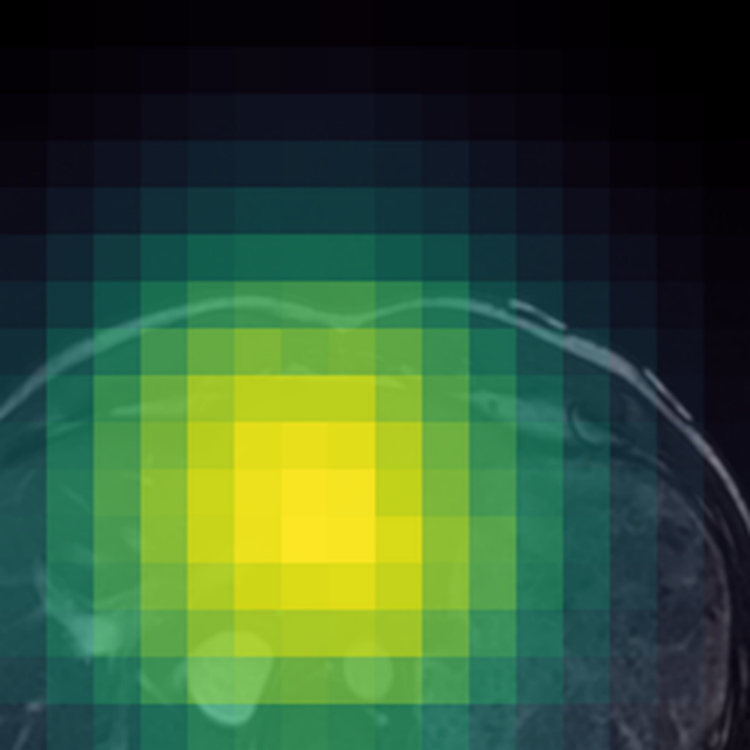}\hfill
        \includegraphics[width=0.24\linewidth]{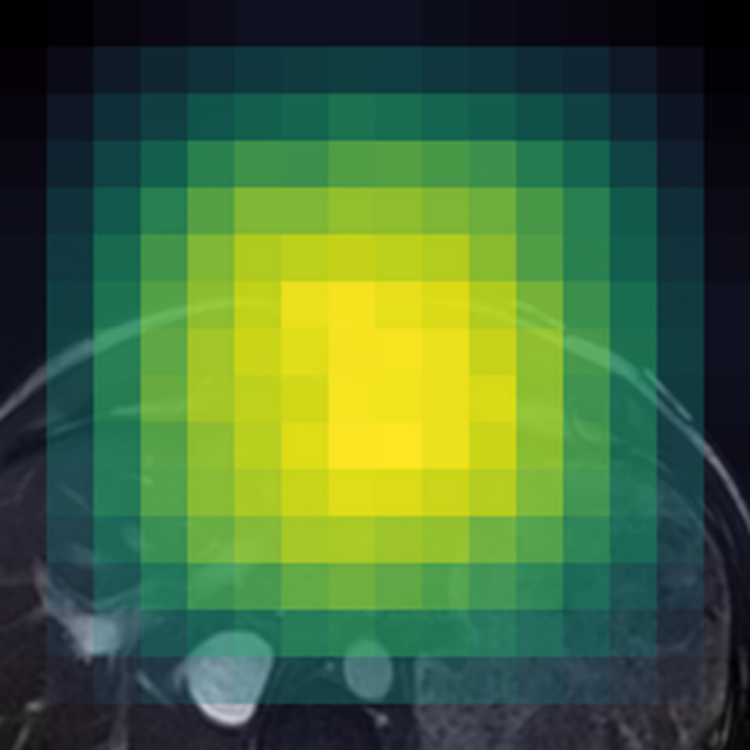}\hfill
        \includegraphics[width=0.24\linewidth]{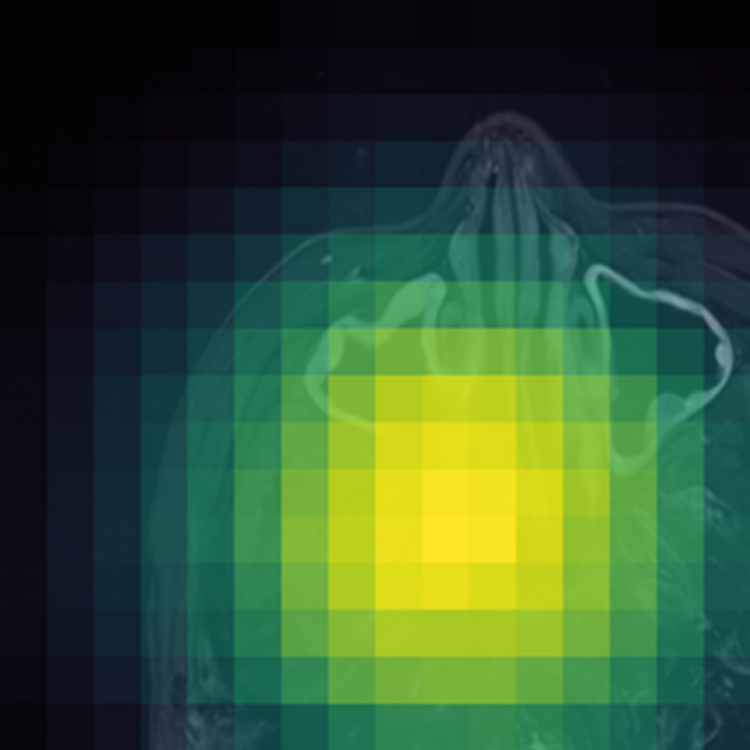}\hfill
        \includegraphics[width=0.24\linewidth]{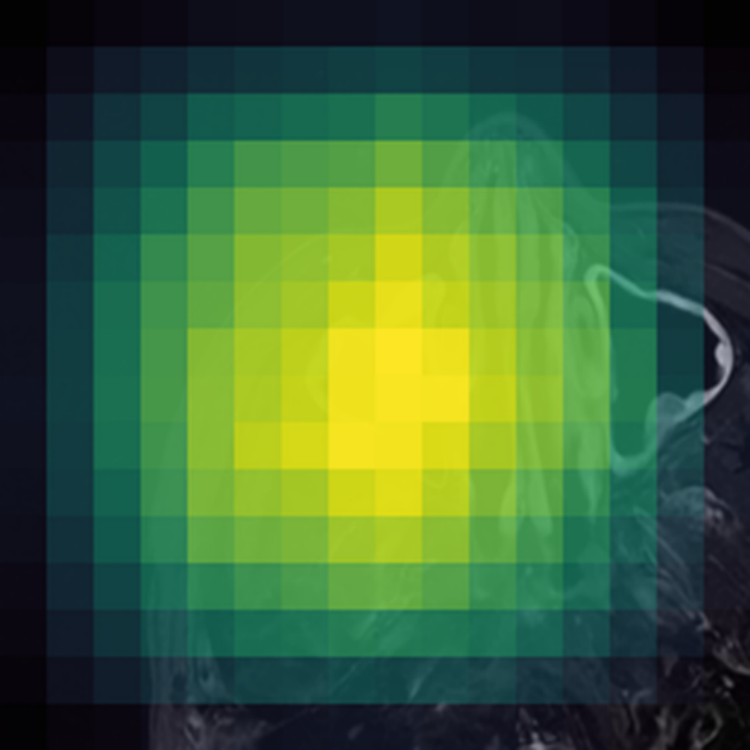}\\[4pt]
        \begin{tabular}{p{0.24\linewidth}p{0.24\linewidth}p{0.24\linewidth}p{0.24\linewidth}}
        \centering \small Anatomically Guided & \centering \small Uniform Prior & \centering \small Anatomically Guided & \centering \small Uniform Prior
    \end{tabular}
    \end{minipage}
    
    \vspace{6pt} 
    \caption{(Left) \textbf{Content-Aware Cropping.} Comparison of valid (green) and filtered out (red) crop regions. (Right) \textbf{Anatomically-Guided Masking.} Each pair shows the average of 2000 mask samplings of our anatomically-guided masking against the blockwise masking with uniform prior strategy (DINOv2~\cite{oquab2023dinov2}).}
    \label{fig:masking}
\end{figure}

%% file: datasets/table.tex
\newcommand{\badge}[1]{%
  \colorbox{gray!10}{\scriptsize\sffamily\bfseries #1}%
}
\newcommand{\taskid}[1]{{\textsf{\textbf{#1}}}}

\begin{table}[!t]
\centering
\caption{\textbf{Evaluation Benchmark.} The tasks are split into 7 categories: Anatomy (A), Oncology (O), Musculoskeletal (M), Emergency (E), Infectious (I), Neurodegenerative (N), Multi-Finding Screening (F). ${\dagger}$: Development task.}
\renewcommand{\arraystretch}{0.0}
\fontsize{8}{9.6}\selectfont
\begin{tabularx}{\textwidth}{@{} c c p{4em} X @{}}
\toprule
\taskid{A1} & ${\dagger}$ & \badge{2D/3D} & \textbf{CT Organ Recognition} in CT scans from TotalSegmentator~\cite{totalsegmentator} dataset. Note that we used 104 organ labels compared to 54 in \cite{dancette2025curia}. \\
\taskid{A2} & ${\dagger}$ & \badge{2D/3D} & \textbf{MRI Organ Recognition} from TotalSegmentator MRI~\cite{d2024totalsegmentatormri} dataset. \\
\taskid{A3} & ${\dagger}$ & \badge{2D/3D} & \textbf{Neuroimaging Age Estimation} from T1-weighted MRI (IXI dataset~\cite{ixidataset}). \\
\midrule
\taskid{O1} & ${\dagger}$ & \badge{2D/3D} & \textbf{Kidney Lesion Malignancy} classification in CT (KITS23~\cite{kits23} dataset). \\
\taskid{O2} & & \badge{2D/3D} & \textbf{Lung Nodule Malignancy} classification in CT (LUNA16 dataset~\cite{setio2017luna16}). \\
\taskid{O3} & & \badge{2D} & \textbf{Tumor Anatomical Localization} prediction in CT (DeepLesion~\cite{yan2018deeplesion}). \\
\taskid{O4} & & \badge{3D}  & \textbf{Kidney Cancer Survival} in contrast-enhanced CT from TCIA\cite{clark2013tcia} cohort. \\
\midrule
\taskid{M1} & ${\dagger}$ & \badge{2D}  & \textbf{Subarticular Stenosis} severity classification in axial T2WI~\cite{rsna-2024-lumbar-spine-degenerative-classification}. \\
\taskid{M2} & & \badge{2D}  & \textbf{Foraminal Narrowing} severity classification in sagittal T1WI~\cite{rsna-2024-lumbar-spine-degenerative-classification}. \\
\taskid{M3} & & \badge{2D}  & \textbf{Spinal Canal Stenosis} severity classification in sagittal T2WI \& STIR~\cite{rsna-2024-lumbar-spine-degenerative-classification}. \\
\taskid{M4} & & \badge{3D}  & \textbf{Anterior Cruciate Ligament (ACL) Tear} severity classification - injury, complete rupture, or absence - in knee MRI exams~\cite{vstajduhar2017semi}. \\
\midrule
\taskid{E1} & ${\dagger}$ & \badge{2D/3D}  & \textbf{Intracranial Hemorrhage} detection in cranial CT (RSNA 2019~\cite{flanders2020construction}). \\
\taskid{E2} & ${\dagger}$ & \badge{2D/3D}   & \textbf{Abdominal Trauma} detection in contrast CT (RSNA 2023 challenge~\cite{rsna-2023-abdominal-trauma-detection}). \\
\taskid{E3} & ${\dagger}$ & \badge{2D/3D}  & \textbf{Myocardial Infarction} signs detection in cardiac MRI (EMIDEC~\cite{emidec}). \\
\taskid{E4} & ${\dagger}$ & \badge{2D}  & \textbf{Stroke}-resulting brain lesions in T1-weighted MRI (ATLAS R2.0~\cite{atlasdataset}). \\
\midrule
\taskid{I} & ${\dagger}$  & \badge{2D} & \textbf{Pulmonary Infections} classification (COVIDx CT~\cite{Gunraj2022}). \\
\midrule
\taskid{N} & & \badge{3D} & \textbf{Alzheimer's Disease} detection in brain MRIs (Oasis-1 dataset~\cite{marcus2007open}). \\
\midrule
\taskid{F} &  & \badge{3D}  & \textbf{Findings Detection} of around 200 clinical entities in images from the Stanford Merlin Abdominal CT Dataset~\cite{blankemeier2024merlin} extracted with RATE~\cite{agrawal2025pillar}. \\
\bottomrule
\end{tabularx}
\label{tab:tasks}
\end{table}

%% file: tables/ablation.tex
\newcommand{\stagebox}[1]{\makebox[4.8em][l]{#1}}
\begin{table}[t]
    \centering
    \caption{\textbf{Ablation Study.} 
    Training collapses when lowering the resolution to 256. 
    Our refinements stabilize convergence on radiological data and consistently improve performance while facilitating the expansion to larger ViTs.
    }
    \fontsize{8}{9.6}\selectfont
    \begin{tabular}{l | l | c c c c c c c c c c}
        \toprule
         \textbf{Strategy} & \textbf{Average} & \shortstack{A1 \\ \scriptsize acc} & \shortstack{A2 \\ \scriptsize acc} & \shortstack{A3 \\ \scriptsize r2} & \shortstack{O1 \\ \scriptsize AUC} & \shortstack{M1 \\ \scriptsize AUC} & \shortstack{E1 \\ \scriptsize AUC} & \shortstack{E2 \\ \scriptsize AUC} & \shortstack{E3 \\ \scriptsize AUC} & \shortstack{E4 \\ \scriptsize AUC} & \shortstack{I \\ \scriptsize bacc}\\
         \midrule
         Curia B (512) & 84.8 & 85.4 & 82.3 &  75.8 & 74.5 & 87.8 & 93.7 & 82.6 & 84.7 & 89.5 & 91.5  	 \\	
         \midrule
         ViT-B (256) & 46.9 \down{37.9} & 22.3 & 17.7 & 0.7 & 49.2 & 68.8 & 69.9 & 64.6 & 64.6 & 74.0 & 37.2 \\
         $+$ Cropping & 81.5 \up{34.6} & 85.4 & 81.4 & 64.8 & 71.0 & 87.2 & 92.1 & 78.8 & 86.8 & 87.1 & 80.5 \\
         $-$ KoLeo & 83.1 \up{1.6} & 84.6 & 81.4 & 74.8 & 70.3 & 87.2 & 93.3 & 82.7 & 82.4 & 87.3 & 87.1 \\
         $+$ SigReg & 83.1 {\scriptsize $=$} & 84.7 & 81.3 & 75.2 & 70.4 & 87.3 & 92.9 & 83.2 & 83.4 & 85.9 & 86.9 \\
         $+$ Masking & 83.3 \up{0.2} & 84.2 & 78.2 & 75.3 & 67.9 & 87.1 & 93.1 & 84.1 & 88.7 & 87.5 & 87.2 \\
         $+$ LR $10^{-3}$ & 83.8 \up{0.5} & 85.1 & 81.7 & 76.0 & 70.6 & 86.7 & 93.0 & 84.7 & 84.8 & 88.0 & 87.4 \\
         $+$ Interp. 512 & 84.9 \up{1.1} & 86.7 & 84.2 & 76.0 & 75.5 & 87.2 & 92.6 & 85.7 & 84.4 & 88.3 & 88.2 \\
         \stagebox{$+$ F-T 512} $=$ Curia-2 B & 85.1 \up{0.2} & 87.3 & 85.2 & 75.4 & 75.3 & 87.3 & 92.6 & 82.7 & 89.0 & 88.8 & 87.3 \\
         \midrule
         \stagebox{$+$ ViT-L} $=$ Curia-2 L & 88.1 \up{3.0} & 91.4 & 89.8 & 77.2 & 84.7 & 87.9 & 93.3 & 87.4 & 89.1 & 89.9 & 90.4 \\
         \stagebox{$+$ ViT-g} $=$ Curia-2 g & 88.7 \up{0.6} & 93.6 & 90.6 & 79.0 & 84.7 & 88.1 & 94.0 & 86.9 & 87.6 & 90.7 & 92.3 \\
         \bottomrule
    \end{tabular}
    \label{tab:ablation}
\end{table}

%% file: figures/similarity-v3.tex
\begin{figure}[t!] 
\centering
    \setlength{\tabcolsep}{2pt} 
    \begin{tabular}{m{0.02\textwidth} m{0.23\textwidth} | m{0.23\textwidth} m{0.23\textwidth} m{0.23\textwidth}}
        \rotatebox{90}{Knee} & 
        \includegraphics[width=0.23\textwidth]{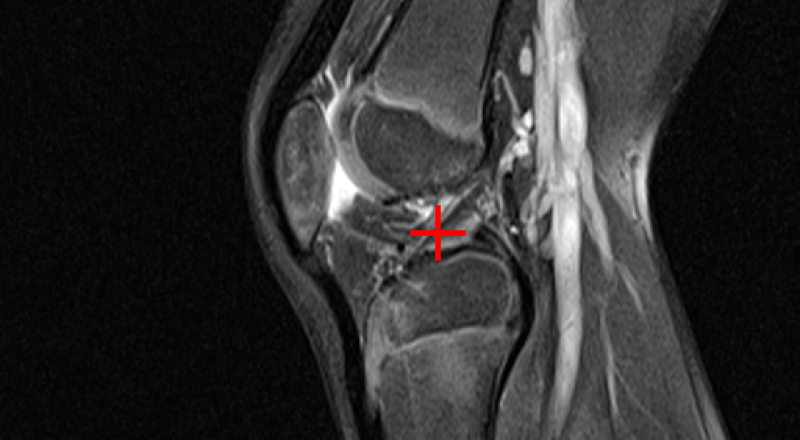} &
        \includegraphics[width=0.23\textwidth]{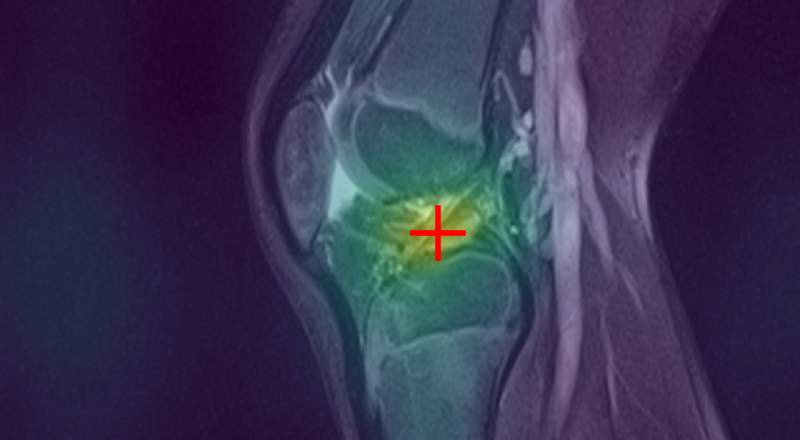} &
        \includegraphics[width=0.23\textwidth]{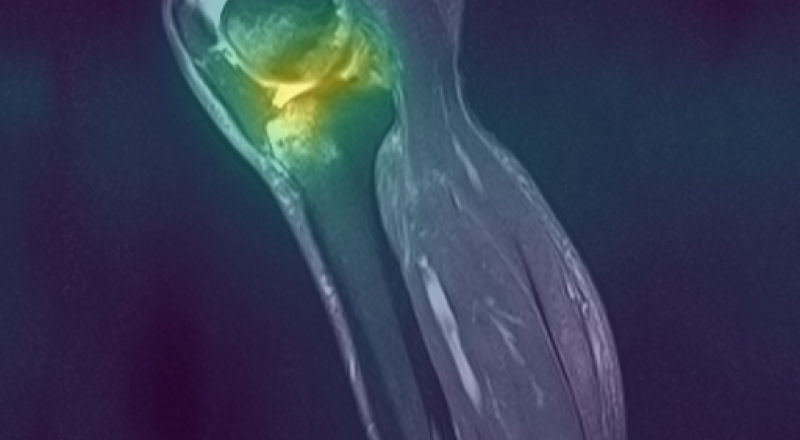} &
        \includegraphics[width=0.23\textwidth]{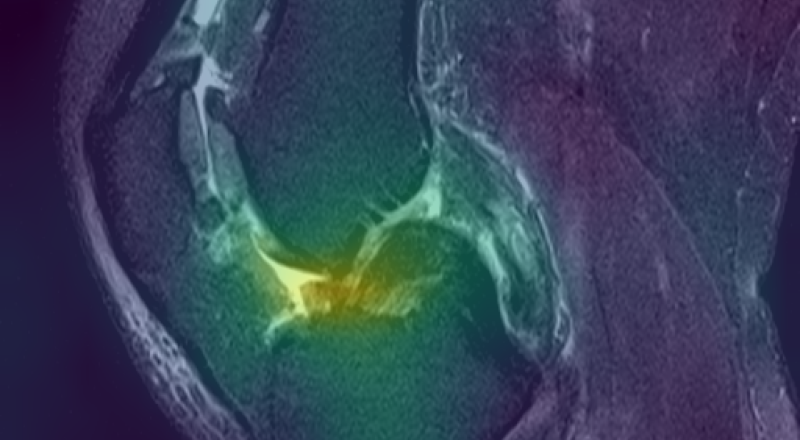} \\
        
        \rotatebox{90}{Abdomen} & 
        \includegraphics[width=0.23\textwidth]{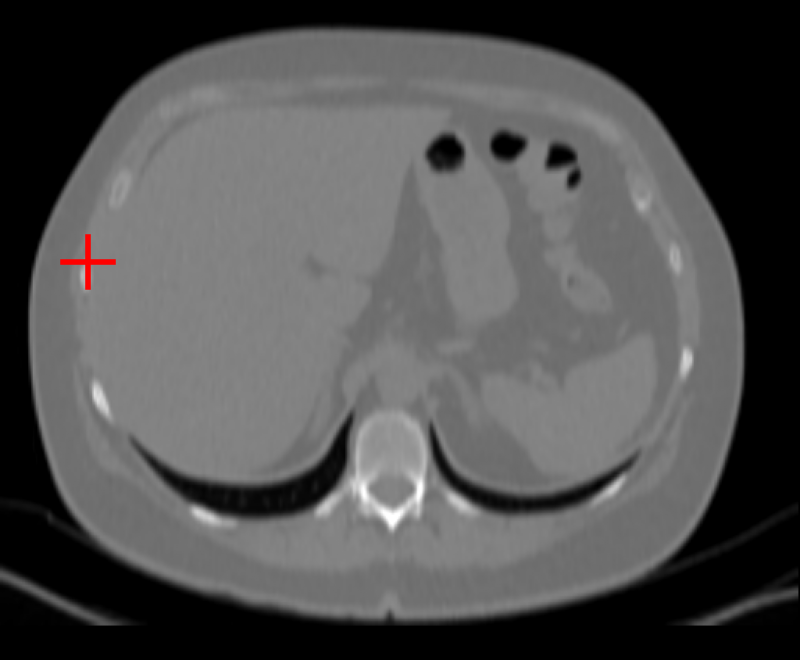} &
        \includegraphics[width=0.23\textwidth]{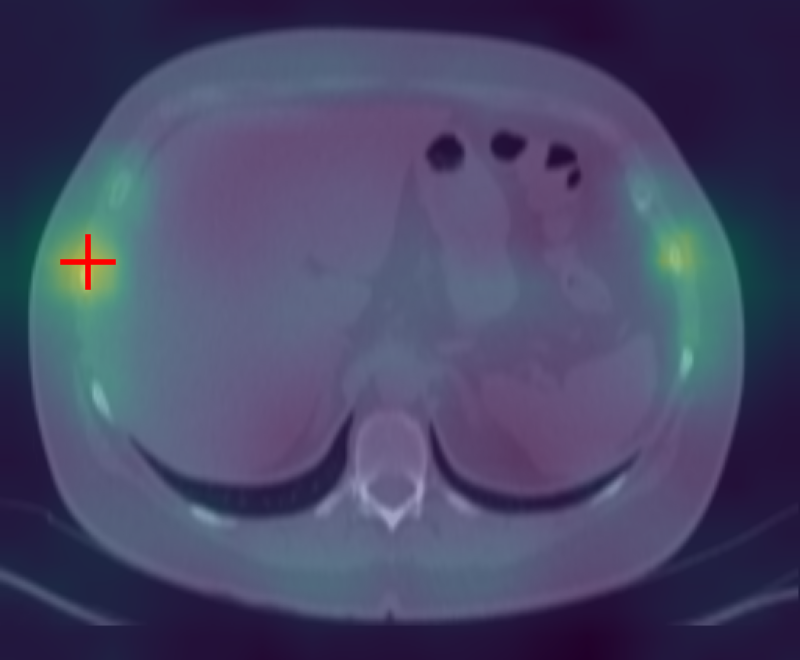} &
        \includegraphics[width=0.23\textwidth]{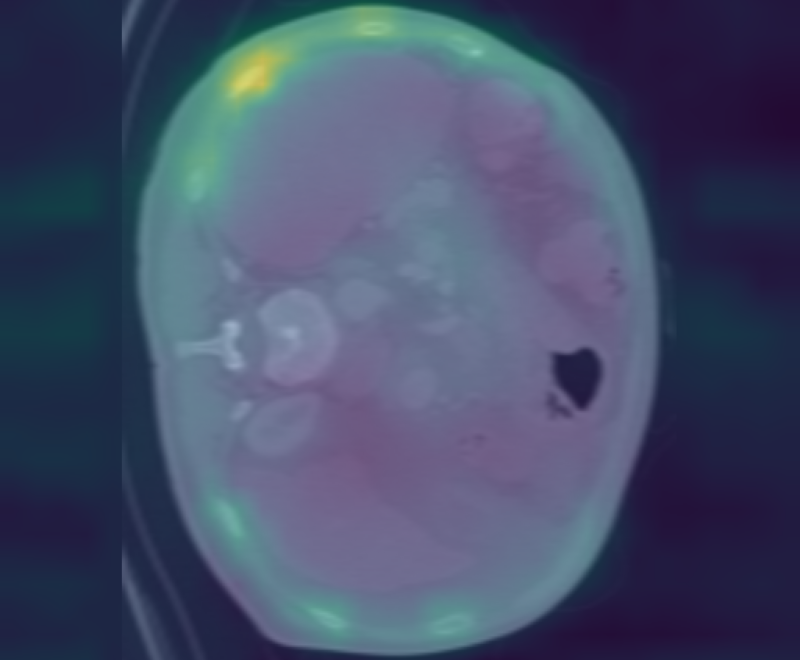} &
        \includegraphics[width=0.23\textwidth]{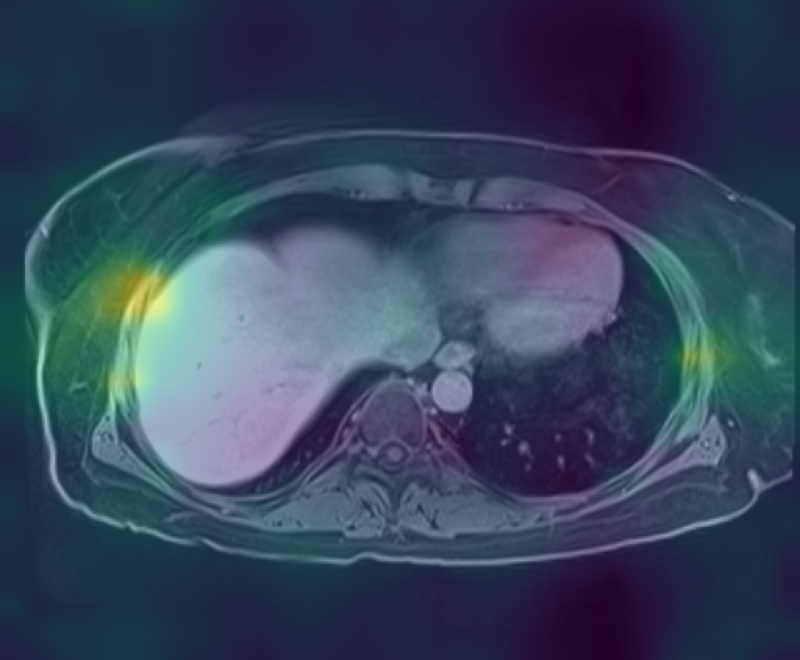} \\
        
        & 
        \multicolumn{1}{p{0.23\textwidth}}{\centering \small Patch Query \textcolor{red}{$\bm{+}$}} &
        \multicolumn{3}{p{0.69\textwidth}}{\centering Cosine Similarity Maps w/ Patch Query}
    \end{tabular}
    \caption{\textbf{Dense Features.} Cosine similarity maps between a query patch \textcolor{red}{$\bm{+}$} and all patches in multiple images. 
    Curia-2 g demonstrates strong semantic understanding of anatomical structures and cross-modality alignment capabilities, mapping structures between CT and MRI domains, even under rotations.
    }
    \label{fig:sim_maps}
\end{figure}

%% file: tables/2dresults.tex
\begin{table}[t]
    \centering

    \caption{\textbf{2D Track Results.} Curia-2 outperforms 2D FMs, including MedImageInsight (MI2), BioMedCLIP (BMCLIP), MedGemma (MGemma) and Curia.}
\fontsize{8}{9.6}\selectfont

\begin{tabular}{l|c|cccccccccccccc}
\toprule
\textbf{Model} & \textbf{Avg.} & \shortstack{A1 \\ \scriptsize acc} & 
\shortstack{A2 \\ \scriptsize acc} & 
\shortstack{A3 \\ \scriptsize r2} & 
\shortstack{O1 \\ \scriptsize AUC} & 
\shortstack{O2 \\ \scriptsize AUC} & 
\shortstack{O3 \\ \scriptsize bacc} & 
\shortstack{M1 \\ \scriptsize AUC} & 
\shortstack{M2 \\ \scriptsize AUC} & 
\shortstack{M3 \\ \scriptsize AUC} & 
\shortstack{E1 \\ \scriptsize AUC} & 
\shortstack{E2 \\ \scriptsize AUC} & 
\shortstack{E3 \\ \scriptsize AUC} & 
\shortstack{E4 \\ \scriptsize AUC} & 
\shortstack{I \\ \scriptsize bacc} \\
\midrule
MI2 & 84.4 & 75.8 & 63.2 & 72.4 & 68.1 & \textbf{92.9} & 88.9 & 85.6 & 86.3 & 93.0 & 90.1 & \textbf{93.2} & \textbf{94.0} & 88.5 & 90.0 \\
BMCLIP & 79.1 & 66.7 & 63.2 & 69.4 & 62.7 & 84.8 & 86.5 & 83.9 & 84.5 & 92.4 & 87.8 & 79.2 & 71.5 & 85.1 & 89.3 \\
MGemma & 80.1 & 67.6 & 63.4 & 57.8 & 63.8 & \underline{91.2} & 89.9 & 83.6 & 84.7 & 91.0 & 87.7 & 85.1 & 77.0 & 88.7 & 90.4 \\
Curia B & 86.2 & 85.4 & 82.3 & 75.8 & 74.4 & 86.6 & \underline{91.9} & 87.8 & 86.4 & \textbf{94.6} & \underline{93.7} & 82.6 & 84.7 & 89.5 & 91.5 \\
Curia L & \underline{87.9} & 88.7 & 89.1 & 75.6 & \underline{80.0} & 84.8 & 91.7 & 87.4 & 86.3 & 93.9 & \underline{93.7} & 87.1 & \underline{89.1} & 89.6 & \textbf{93.4} \\
\midrule
\rowcolor{blue!10} Curia-2 B & 86.8 & 87.2 & 85.2 & 75.4 & 75.3 & 89.6 & \textbf{93.5} & 87.3 & \underline{86.9} & \underline{94.5} & 92.6 & 82.7 & 89.0 & 88.8 & 87.3 \\
\rowcolor{blue!10} Curia-2 L & \textbf{88.5} & \underline{91.4} & \underline{89.8} & \underline{77.2} & \textbf{84.7} & 84.0 & 91.8 & \underline{87.9} & \textbf{87.4} & \underline{94.5} & 93.3 & \underline{87.4} & \underline{89.1} & \underline{89.9} & 90.4 \\
\rowcolor{blue!10} Curia-2 g & \textbf{88.5} & \textbf{93.6} & \textbf{90.6} & \textbf{79.0} & \textbf{84.7} & 80.1 & 90.6 & \textbf{88.1} & 86.6 & \underline{94.5} & \textbf{94.0} & 86.9 & 87.6 & \textbf{90.7} & \underline{92.3} \\
\bottomrule
\end{tabular}

    \label{tab:2dresults}
\end{table}

%% file: tables/3dresults.tex
\begin{table}[t]
    \centering

    \caption{\textbf{3D Track results.} Curia-2 scores the best results on this generalist benchmark against both 3D volumetric FMs and 2D slice-based FMs.}
\fontsize{8}{9.6}\selectfont

\begin{tabular}{l|c|cccccccccccc}
\toprule
\textbf{Model} & \textbf{Avg}. & \shortstack{A1 \\ \scriptsize acc} & \shortstack{A2 \\ \scriptsize acc} & \shortstack{A3 \\ \scriptsize r2} & \shortstack{O1 \\ \scriptsize AUC} & \shortstack{O2 \\ \scriptsize AUC} & \shortstack{O4 \\ \scriptsize cindex} & \shortstack{M4 \\ \scriptsize AUC} & \shortstack{E1 \\ \scriptsize AUC} & \shortstack{E2 \\ \scriptsize AUC} & \shortstack{E3 \\ \scriptsize AUC} & \shortstack{N \\ \scriptsize bacc} & \shortstack{F \\ \scriptsize acc} \\
\midrule
CT-CLIP & 51.5 & 38.1 & 20.9 & 22.0 & 63.8 & 56.8 & 58.1 & 43.5 & 69.3 & 42.6 & 69.7 & 69.9 & 63.4 \\
Merlin & 65.4 & 45.4 & 38.5 & 53.3 & 55.7 & 64.9 & 66.3 & 50.4 & 85.1 & 92.5 & 75.6 & 76.0 & \underline{81.2} \\
Pillar-0 & 71.5 & 50.7 & 40.0 & 53.5 & 70.2 & 84.9 & 68.1 & 69.2 & 92.4 & \textbf{98.2} & 56.9 & 91.7 & \textbf{82.4} \\
\midrule
MI2 & 83.6 & 86.3 & 83.0 & 80.4 & 79.8 & \textbf{89.8} & 59.3 & 71.2 & 91.5 & 94.4 & 94.6 & \underline{94.4} & 79.0 \\
BMCLIP & 78.3 & 77.8 & 71.9 & 85.0 & 64.1 & 81.3 & 62.3 & 72.8 & 89.2 & 90.1 & 79.3 & 91.7 & 74.1 \\
MGemma & 81.8 & 76.2 & 79.5 & 77.7 & 77.9 & \underline{88.7} & \underline{75.3} & 70.1 & 87.9 & 94.0 & 89.7 & 91.2 & 72.7 \\
Curia B & 86.0 & 88.2 & 92.7 & 87.8 & 82.6 & 77.8 & 74.3 & 78.2 & \textbf{95.1} & 93.5 & 93.8 & 89.0 & 78.8 \\
Curia L & 84.5 & 89.4 & \underline{95.7} & 86.7 & 84.0 & 77.6 & 60.6 & 74.4 & 94.1 & 91.9 & 87.9 & 92.4 & 79.2 \\
\midrule
\rowcolor{blue!10} Curia-2 B & 87.7 & 90.9 & 95.0 & 87.4 & 84.5 & 78.5 & 72.5 & \underline{81.1} & 93.7 & \underline{94.7} & \textbf{99.2} & \textbf{94.7} & 80.2 \\
\rowcolor{blue!10} Curia-2 L & \textbf{88.6} & \underline{93.2} & \underline{95.7} & \underline{88.9} & \underline{87.1} & 79.3 & \textbf{75.5} & \textbf{84.9} & 94.2 & 93.2 & \underline{98.6} & 92.6 & 79.7 \\
\rowcolor{blue!10} Curia-2 g & \underline{87.8} & \textbf{94.2} & \textbf{96.5} & \textbf{89.7} & \textbf{87.9} & 75.3 & 74.6 & 79.4 & \underline{94.6} & 93.8 & 96.1 & 92.8 & 79.3 \\
\bottomrule
\end{tabular}

    \label{tab:3dresults}
\end{table}

%% file: main.bib
@article{oquab2023dinov2,
title={{DINO}v2: Learning Robust Visual Features without Supervision},
author={Maxime Oquab and Timoth{\'e}e Darcet and Th{\'e}o Moutakanni and Huy V. Vo and Marc Szafraniec and Vasil Khalidov and Pierre Fernandez and Daniel HAZIZA and Francisco Massa and Alaaeldin El-Nouby and Mido Assran and Nicolas Ballas and Wojciech Galuba and Russell Howes and Po-Yao Huang and Shang-Wen Li and Ishan Misra and Michael Rabbat and Vasu Sharma and Gabriel Synnaeve and Hu Xu and Herve Jegou and Julien Mairal and Patrick Labatut and Armand Joulin and Piotr Bojanowski},
journal={Transactions on Machine Learning Research},
issn={2835-8856},
year={2024}
}

@article{totalsegmentator,
  title     = {TotalSegmentator: Robust Segmentation of 104 Anatomic Structures in CT Images},
  volume    = {5},
  ISSN      = {2638-6100},
  number    = {5},
  journal   = {Radiology: Artificial Intelligence},
  publisher = {Radiological Society of North America (RSNA)},
  author    = {Wasserthal, Jakob and Breit, Hanns-Christian and Meyer, Manfred T. and Pradella, Maurice and Hinck, Daniel and Sauter, Alexander W. and Heye, Tobias and Boll, Daniel T. and Cyriac, Joshy and Yang, Shan and Bach, Michael and Segeroth, Martin},
  year      = {2023},
  month     = sep
}

@book{kits23,
  title     = {Deep Learning in Medical Image Analysis and Multimodal Learning for Clinical Decision Support: Third International Workshop, DLMIA 2017, and 7th International Workshop, ML-CDS 2017, Held in Conjunction with MICCAI 2017, Qu{\'e}bec City, QC, Canada, September 14, Proceedings},
  author    = {Cardoso, M. Jorge and Arbel, Tal and Carneiro, Gustavo and Syeda-Mahmood, Tanveer and Tavares, Jo{\~a}o Manuel RS and Moradi, Mehdi and Bradley, Andrew and Greenspan, Hayit and Papa, Jo{\~a}o Paulo and Madabhushi, Anant and others},
  volume    = {10553},
  year      = {2017},
  publisher = {Springer}
}

@article{d2024totalsegmentatormri,
  title={TotalSegmentator MRI: Robust sequence-independent segmentation of multiple anatomic structures in MRI},
  author={D'Antonoli, Tugba Akinci and Berger, Lucas K and Indrakanti, Ashraya K and Vishwanathan, Nathan and Wei{\ss}, Jakob and Jung, Matthias and Berkarda, Zeynep and Rau, Alexander and Reisert, Marco and K{\"u}stner, Thomas and others},
  journal={arXiv preprint arXiv:2405.19492},
  year={2024}
}

@article{setio2017luna16,
  title     = {Validation, comparison, and combination of algorithms for automatic detection of pulmonary nodules in computed tomography images: the LUNA16 challenge},
  author    = {Setio, Arnaud Arindra Adiyoso and Traverso, Alberto and De Bel, Thomas and Berens, Moira S. N. and van den Bogaard, Cas and Cerello, Piergiorgio and Chen, Hao and Dou, Qi and Fantacci, Maria Evelina and Geurts, Bram and others},
  journal   = {Medical Image Analysis},
  volume    = {42},
  pages     = {1--13},
  year      = {2017},
  publisher = {Elsevier}
}

@article{vstajduhar2017semi,
  title   = {Semi-automated detection of anterior cruciate ligament injury from MRI},
  author  = {{\v{S}}tajduhar, Ivan and Mamula, Mihaela and Mileti{\'c}, Damir and Uenal, Goezde},
  journal = {Computer Methods and Programs in Biomedicine},
  volume  = {140},
  pages   = {151--164},
  year    = {2017},
  publisher = {Elsevier}
}

@article{marcus2007open,
  title     = {Open Access Series of Imaging Studies (OASIS): cross-sectional MRI data in young, middle aged, nondemented, and demented older adults},
  author    = {Marcus, Daniel S. and Wang, Tracy H. and Parker, Jamie and Csernansky, John G. and Morris, John C. and Buckner, Randy L.},
  journal   = {Journal of Cognitive Neuroscience},
  volume    = {19},
  number    = {9},
  pages     = {1498--1507},
  year      = {2007},
  publisher = {MIT Press}
}

@article{zhang2025BioMedCLIP,
  title     = {BioMedCLIP: A Multimodal Biomedical Foundation Model Trained from Fifteen Million Image--Text Pairs},
  author    = {Zhang, Sheng and Xu, Yanbo and Usuyama, Naoto and Xu, Hanwen and Bagga, Jaspreet and Tinn, Robert and Preston, Sam and Rao, Rajesh and Wei, Mu and Valluri, Naveen and Wong, Cliff and Tupini, Andrea and Wang, Yu and Mazzola, Matt and Shukla, Swadheen and Liden, Lars and Gao, Jianfeng and Crabtree, Angela and Piening, Brian and Bifulco, Carlo and Lungren, Matthew P. and Naumann, Tristan and Wang, Sheng and Poon, Hoifung},
  year      = {2025},
  month     = jan,
  journal   = {NEJM AI},
  volume    = {2},
  number    = {1},
  publisher = {Massachusetts Medical Society},
}

@misc{codella2024MedImageInsight,
  title         = {MedImageInsight: An Open-Source Embedding Model for General Domain Medical Imaging},
  shorttitle    = {MedImageInsight},
  author        = {Codella, Noel C. F. and Jin, Ying and Jain, Shrey and Gu, Yu and Lee, Ho Hin and Ben Abacha, Asma and Santamaria-Pang, Alberto and Guyman, Will and Sangani, Naiteek and Zhang, Sheng and Poon, Hoifung and Hyland, Stephanie and Bannur, Shruthi and Alvarez-Valle, Javier and Li, Xue and Garrett, John and McMillan, Alan and Rajguru, Gaurav and Maddi, Madhu and Vijayrania, Nilesh and Bhimai, Rehaan and Mecklenburg, Nick and Jain, Rupal and Holstein, Daniel and Gaur, Naveen and Aski, Vijay and Hwang, Jenq-Neng and Lin, Thomas and Tarapov, Ivan and Lungren, Matthew and Wei, Mu},
  year          = {2024},
  month         = oct,
  number        = {arXiv:2410.06542},
  eprint        = {2410.06542},
  primaryclass  = {eess},
  publisher     = {arXiv},
  archiveprefix = {arXiv}
}

@misc{ixidataset,
  title = {IXI Dataset},
  url   = {https://brain-development.org/ixi-dataset}
}

@article{atlasdataset,
  title={A large, curated, open-source stroke neuroimaging dataset to improve lesion segmentation algorithms},
  author={Liew, Sook-Lei and Lo, Bethany P and Donnelly, Miranda R and Zavaliangos-Petropulu, Artemis and Jeong, Jessica N and Barisano, Giuseppe and Hutton, Alexandre and Simon, Julia P and Juliano, Julia M and Suri, Anisha and others},
  journal={Scientific data},
  volume={9},
  number={1},
  pages={320},
  year={2022},
  publisher={Nature Publishing Group UK London}
}

@article{flanders2020construction,
  title     = {Construction of a machine learning dataset through collaboration: the RSNA 2019 brain CT hemorrhage challenge},
  author    = {Flanders, Adam E. and Prevedello, Luciano M. and Shih, George and Halabi, Safwan S. and Kalpathy-Cramer, Jayashree and Ball, Robyn and Mongan, John T. and Stein, Anouk and Kitamura, Felipe C. and Lungren, Matthew P. and others},
  journal   = {Radiology: Artificial Intelligence},
  volume    = {2},
  number    = {3},
  pages     = {e190211},
  year      = {2020},
  publisher = {Radiological Society of North America}
}

@article{yan2018deeplesion,
  title     = {DeepLesion: automated mining of large-scale lesion annotations and universal lesion detection with deep learning},
  author    = {Yan, Ke and Wang, Xiaosong and Lu, Le and Summers, Ronald M.},
  journal   = {Journal of Medical Imaging},
  volume    = {5},
  number    = {3},
  pages     = {036501--036501},
  year      = {2018},
  publisher = {SPIE}
}

@article{clark2013tcia,
  title     = {The Cancer Imaging Archive (TCIA): maintaining and operating a public information repository},
  author    = {Clark, Kenneth and Vendt, Bruce and Smith, Kirk and Freymann, John and Kirby, Justin and Koppel, Paul and Moore, Stephen and Phillips, Stanley and Maffitt, David and Pringle, Michael and others},
  journal   = {Journal of Digital Imaging},
  volume    = {26},
  pages     = {1045--1057},
  year      = {2013},
  publisher = {Springer}
}

@article{Gunraj2022,
  author  = {Gunraj, Hayden and Sabri, Ali and Koff, David and Wong, Alexander},
  title   = {COVID-Net CT-2: Enhanced Deep Neural Networks for Detection of COVID-19 From Chest CT Images Through Bigger, More Diverse Learning},
  journal = {Frontiers in Medicine},
  volume  = {8},
  pages   = {729287},
  year    = {2022},
  issn    = {2296-858X}
}

@inproceedings{vit,
  author    = {Dosovitskiy, Alexey and Beyer, Lucas and Kolesnikov, Alexander and Weissenborn, Dirk and Zhai, Xiaohua and Unterthiner, Thomas and Dehghani, Mostafa and Minderer, Matthias and Heigold, Georg and Gelly, Sylvain and Uszkoreit, Jakob and Houlsby, Neil},
  title     = {An Image is Worth 16x16 Words: Transformers for Image Recognition at Scale},
  booktitle = {International Conference on Learning Representations},
  year      = {2021},
editor={ICLR}
}

@misc{rsna-2024-lumbar-spine-degenerative-classification,
  author      = {Tyler Richards and Jason Talbott and Robyn Ball and Errol Colak and Adam Flanders and Felipe Kitamura and John Mongan and Luciano Prevedello and Maryam Vazirabad},
  title       = {RSNA 2024 Lumbar Spine Degenerative Classification},
  year        = {2024},
  howpublished= {\url{https://kaggle.com/competitions/rsna-2024-lumbar-spine-degenerative-classification}}
}

@article{emidec,
  author  = {Lalande, Alain and Chen, Zhihao and Decourselle, Thomas and Qayyum, Abdul and Pommier, Thibaut and Lorgis, Luc and de la Rosa, Ezequiel and Cochet, Alexandre and Cottin, Yves and Ginhac, Dominique and Salomon, Michel and Couturier, Rapha{\"e}l and Meriaudeau, Fabrice},
  year    = {2020},
  month   = {09},
  title   = {EMIDEC: A Database Usable for the Automatic Evaluation of Myocardial Infarction from Delayed-Enhancement Cardiac MRI},
  volume  = {5},
  journal = {Data},
}

@misc{rsna-2023-abdominal-trauma-detection,
  author      = {Errol Colak and Hui-Ming Lin and Robyn Ball and Melissa Davis and Adam Flanders and Sabeena Jalal and Kirti Magudia and Brett Marinelli and Savvas Nicolaou and Luciano Prevedello and Jeff Rudie and George Shih and Maryam Vazirabad and John Mongan},
  title       = {RSNA 2023 Abdominal Trauma Detection},
  year        = {2023},
  howpublished= {\url{https://kaggle.com/competitions/rsna-2023-abdominal-trauma-detection}}
}

@article{dancette2025curia,
  title={Curia: A Multi-Modal Foundation Model for Radiology},
  author={Dancette, Corentin and Khlaut, Julien and Saporta, Antoine and Philippe, Helene and Ferreres, Elodie and Callard, Baptiste and Danielou, Th{\'e}o and Alberge, L{\'e}o and Machado, L{\'e}o and Tordjman, Daniel and others},
  journal={arXiv preprint arXiv:2509.06830},
  year={2025}
}

@article{simeoni2025dinov3,
  title={Dinov3},
  author={Sim{\'e}oni, Oriane and Vo, Huy V and Seitzer, Maximilian and Baldassarre, Federico and Oquab, Maxime and Jose, Cijo and Khalidov, Vasil and Szafraniec, Marc and Yi, Seungeun and Ramamonjisoa, Micha{\"e}l and others},
  journal={arXiv preprint arXiv:2508.10104},
  year={2025}
}

@article{balestriero2025lejepa,
  title={Lejepa: Provable and scalable self-supervised learning without the heuristics},
  author={Balestriero, Randall and LeCun, Yann},
  journal={arXiv preprint arXiv:2511.08544},
  year={2025}
}

@article{agrawal2025pillar,
  title={Pillar-0: A new frontier for radiology foundation models},
  author={Agrawal, Kumar Krishna and Liu, Longchao and Lian, Long and Nercessian, Michael and Harguindeguy, Natalia and Wu, Yufu and Mikhael, Peter and Lin, Gigin and Sequist, Lecia V and Fintelmann, Florian and others},
  journal={arXiv preprint arXiv:2511.17803},
  year={2025}
}

@article{blankemeier2024merlin,
  title={Merlin: A vision language foundation model for 3d computed tomography},
  author={Blankemeier, Louis and Cohen, Joseph Paul and Kumar, Ashwin and Van Veen, Dave and Gardezi, Syed Jamal Safdar and Paschali, Magdalini and Chen, Zhihong and Delbrouck, Jean-Benoit and Reis, Eduardo and Truyts, Cesar and others},
  journal={Research Square},
  pages={rs--3},
  year={2024}
}

@article{sellergren2025medgemma,
  title={Medgemma technical report},
  author={Sellergren, Andrew and Kazemzadeh, Sahar and Jaroensri, Tiam and Kiraly, Atilla and Traverse, Madeleine and Kohlberger, Timo and Xu, Shawn and Jamil, Fayaz and Hughes, C{\'\i}an and Lau, Charles and others},
  journal={arXiv preprint arXiv:2507.05201},
  year={2025}
}

@article{zhou2021ibot,
  title={ibot: Image bert pre-training with online tokenizer},
  author={Zhou, Jinghao and Wei, Chen and Wang, Huiyu and Shen, Wei and Xie, Cihang and Yuille, Alan and Kong, Tao},
  journal={arXiv preprint arXiv:2111.07832},
  year={2021}
}
